\documentclass[runningheads]{llncs}

\usepackage[T1]{fontenc}
\usepackage{graphicx}
\usepackage{booktabs}
\usepackage{multirow}
\usepackage{colortbl}
\usepackage{color}
\usepackage[pagebackref,breaklinks,colorlinks]{hyperref}

\usepackage{wrapfig}
\usepackage{orcidlink}
\usepackage{xr-hyper}
\usepackage{algorithm}
\usepackage{algpseudocode}
\usepackage{amsmath,bm}
\usepackage{bbm}
\usepackage{amsfonts}
\usepackage{caption}
\usepackage{subfig}
\usepackage{siunitx}

\usepackage{graphicx}
\usepackage[table]{xcolor} 
\usepackage{color} 
\definecolor{Gray}{gray}{0.9}

\usepackage[euler]{textgreek}
\usepackage{arydshln}
\usepackage{marvosym}

\usepackage{tikz}
\newcommand*\circled[1]{\tikz[baseline=(char.base)]{
            \node[shape=circle,draw,inner sep=1pt] (char) {#1};}}
            
\begin{document}

\title{Semi-Supervised Few-Shot Adaptation \\ of Vision-Language Models}
\titlerunning{Semi-Supervised Few-Shot Adaptation \\ of Vision-Language Models}
 

%
%

\author{Julio Silva-Rodríguez \and Ender Konukoglu}
\authorrunning{J. Silva-Rodr\'iguez and E. Konukoglu}
\institute{Computer Vision Lab, ETH Zurich, Zurich, Switzerland \\
    \email{jusilva@ethz.ch}}


\maketitle          

\begin{abstract}

Vision-language models (VLMs) pre-trained on large, heterogeneous data sources are becoming increasingly popular, providing rich multi-modal embeddings that enable efficient transfer to new tasks. A particularly relevant application is few-shot adaptation, where only a handful of annotated examples are available to adapt the model through multi-modal linear probes. In medical imaging, specialized VLMs have shown promising performance in zero- and few-shot image classification, which is valuable for mitigating the high cost of expert annotations. However, challenges remain in extremely low-shot regimes: the inherent class imbalances in medical tasks often lead to underrepresented categories, penalizing overall model performance. To address this limitation, we propose leveraging unlabeled data by introducing an efficient semi-supervised solver that propagates text-informed pseudo-labels during few-shot adaptation. The proposed method enables lower-budget annotation pipelines for adapting VLMs, reducing labeling effort by $\geq50\%$ in low-shot regimes. Code is available: \href{https://github.com/jusiro/SS-Text-U}{SS-Text-U}.

\end{abstract}

\section{Introduction}
\label{sec:intro}

\begin{figure*}[t!]
    \begin{center}
        \begin{tabular}{ccccc}

         \multicolumn{3}{c}{
         \includegraphics[width=.56\linewidth]{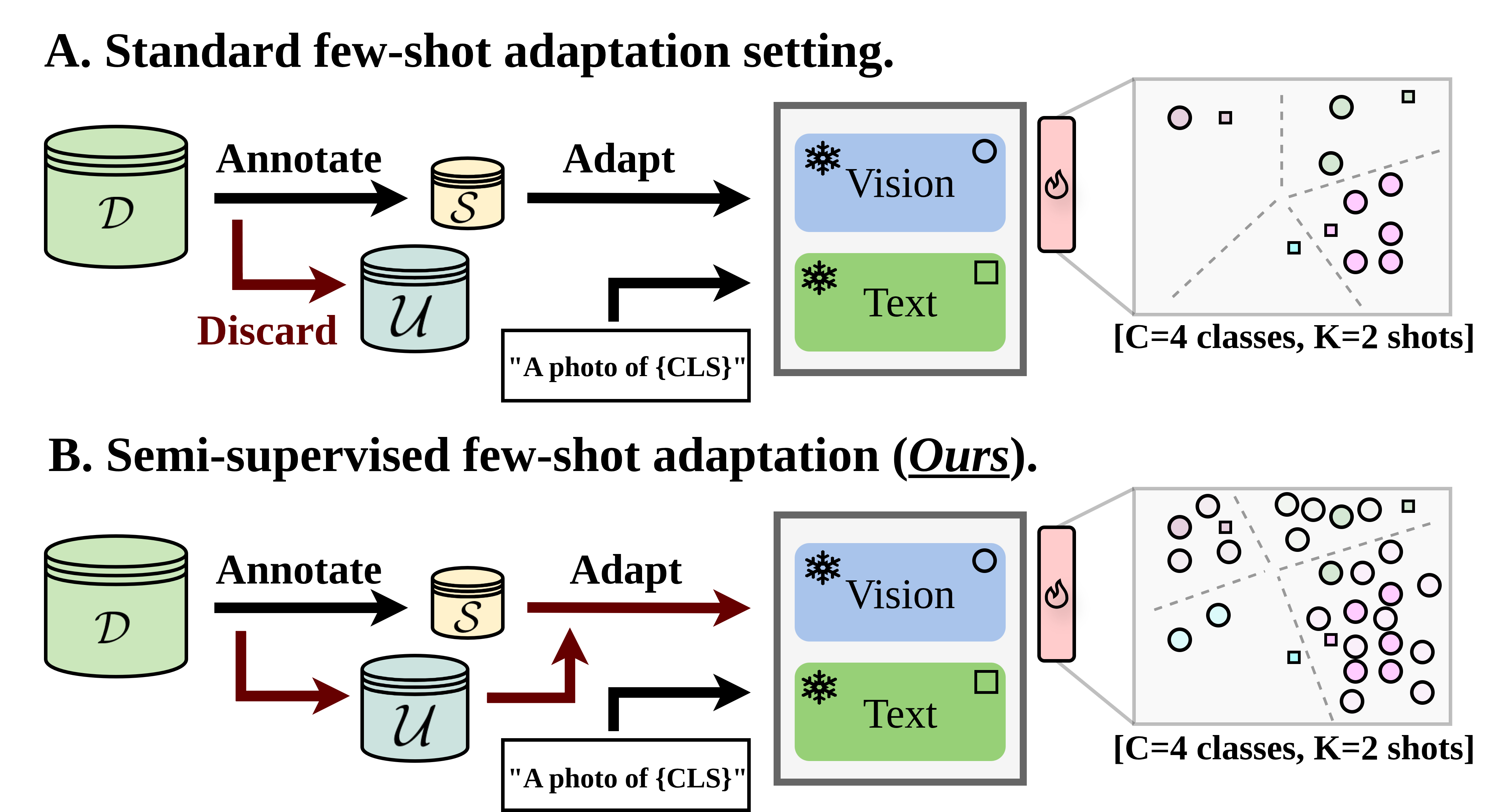}
         }
         &          
         \multicolumn{2}{c}{
         \includegraphics[width=.38\linewidth]{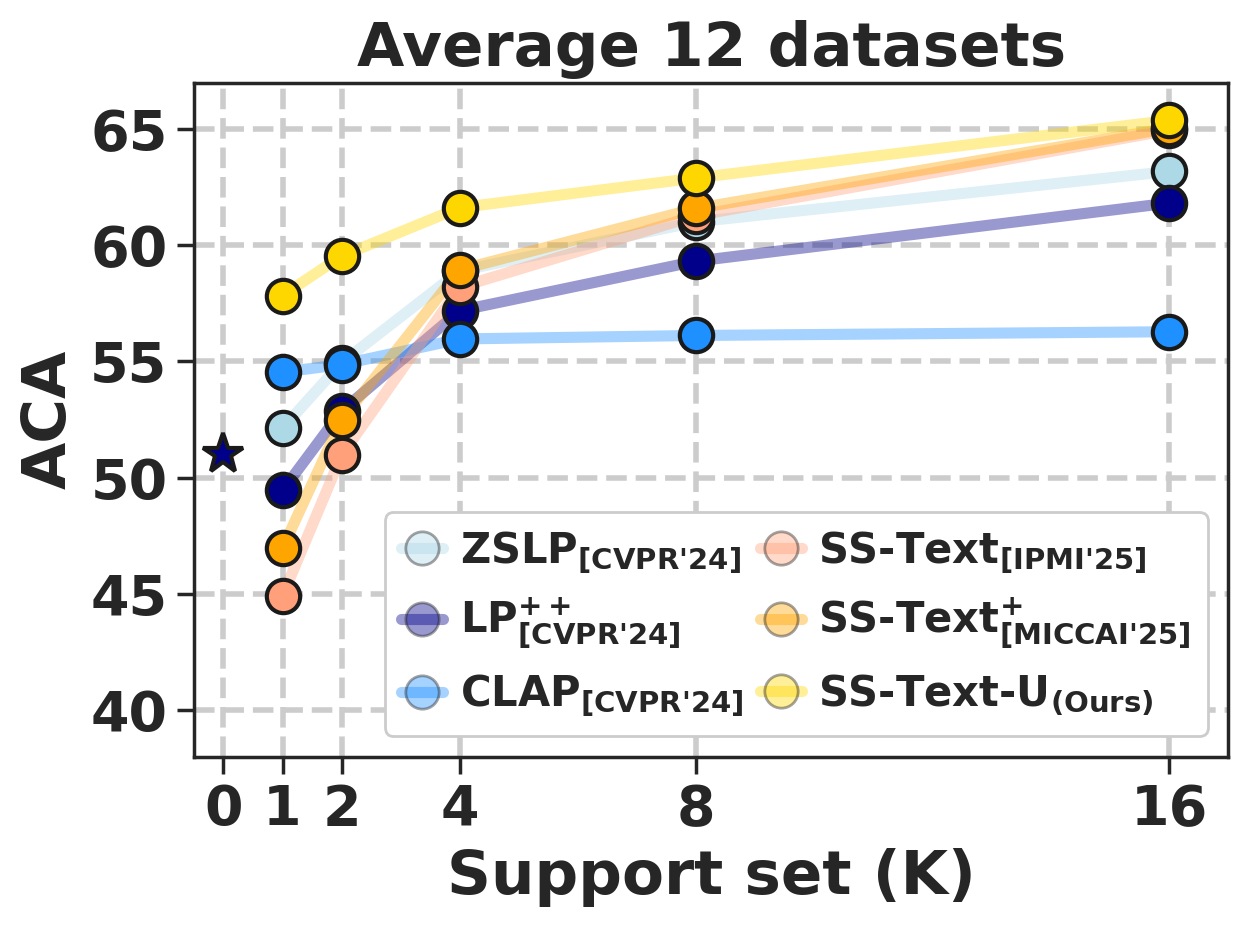}
         } \\

          \multicolumn{3}{c}{\textbf{(a) Proposed pipeline}} & \multicolumn{2}{c}{\textbf{(b) SoTA comparison}} \\

        \end{tabular}
        \caption{\textbf{Semi-supervised few-shot VLMs adaptation}.}
        \label{fig:motivation}
    \end{center}
\end{figure*}

Foundation models, particularly contrastive vision-language models (VLMs), are being increasingly adopted in computer vision. These learn flexible multi-modal embedding representations that can be transferred to downstream tasks, and have shown improved robustness across domain shifts and data-efficient adaptation capabilities \cite{radford2021learning}. The limited performance of generalist VLMs in visually fine-grained, specialized domains such as medical imaging has motivated the development of modality-specific medical VLMs, e.g., for histology~\cite{CONCH}, ophthalmology~\cite{FLAIR}, or radiology~\cite{convirt}. A widely recognized use case for these VLMs in medical image computing is few-shot classification, i.e., using only a small number of labeled images to adapt the model to a new dataset/task (so-called shots), given the high cost of gathering expert-level annotations. First developed in the computer vision community, early work focused on prompt learning techniques \cite{zhou2022coop}, which were later superseded by efficient text-informed linear probes \cite{lp24,clap24}. The latter methods provide similar performance while having orders of magnitude less computational complexity \cite{lp24}. Rapidly, the setting was adopted for adapting medical VLMs \cite{shakeri2024few}. Currently, few-shot adaptation is one of the \textit{de facto} evaluation scenarios for the multi-modal transfer capabilities of newly introduced VLMs \cite{CONCH,ikezogwo2023quiltm,FLAIR,dlilp}. However, this topic is still in its early stages of development, and numerous challenges remain to be explored. For example, as discussed in \cite{sstext25}, medical datasets often exhibit highly imbalanced class distributions, leading to support sets with underrepresented categories and thereby affecting the final model's performance in low-shot regimes. In this work, we propose alleviating this issue by using unlabeled data, specifically via pseudo-labels propagated from textual priors. This scenario is appealing since in any data management pipeline, a sufficient amount of unlabeled data is usually assumed to be gathered, and the constraint typically lies in the annotation stage. Interestingly, despite the clear potential use case, we find this topic largely unexplored in the few-shot literature. While some works have proposed exploiting unlabeled data for adapting VLMs, these typically involve test-time data, e.g., in transduction \cite{zanella2024boosting} or test-time adaptation \cite{shu2022tpt,tta_medvlm}. Considering all of the above, our \textbf{\textit{main contributions}} are: \textbf{\circled{1}}~We introduce semi-supervised few-shot learning (Fig.~\ref{fig:motivation}(a)), a setting to leverage unlabeled data to enable a more annotation-efficient transfer of VLMs; \textbf{\circled{2}}~We propose SS-Text-U. This principled text-informed linear probe simultaneously learns class prototypes and pseudo-labels from few-shot labeled and unlabeled data via an efficient block-wise optimizer and Optimal Transport. Comprehensive experiments on 12 datasets and 3 modality-specialized medical VLMs demonstrate the benefits of our solver over vanilla few-shot adapters (Fig.~\ref{fig:motivation}(b)) to improve predictive performance at lower annotation budgets.

\section{Background}
\label{sec:background}

\subsection{Zero-shot models}
\label{ssec:zeroshot}

Contrastive VLMs following CLIP \cite{radford2021learning} are composed of a visual and a textual encoders, which represent each modality input into an $\ell_{2}$-normalized $D$-dimensional shared feature space, yielding the corresponding visual, ${\mathbf{v}} \in \mathbb{R}^{D \times 1}$, and text, ${\mathbf{t}} \in \mathbb{R}^{D \times 1}$, embeddings. Given a set of $C$ candidate labels, $\mathcal{Y}\in\{1,...,c,...,C\}$, and its prototypes, $\mathbf{W}=(\mathbf{w}_c)_{1 \leq c \leq C}$, with $\mathbf{w}_c \in \mathbb{R}^{D \times 1}$, image class assignments for a particular sample embedding, $\mathbf{v}_i$, can be obtained as:
\begin{align}
\label{eq:probs}
    \phantom{,}p_{i,c}(\mathbf{W})
    = \frac
    {\exp( \mathbf{v}_i^\top \mathbf{w}_{c} / \tau)}
    {\sum_{j \in \mathcal{Y}}\exp( \mathbf{v}^\top \mathbf{w}_j / \tau)},
\end{align}
where $\tau$ is a pre-trained temperature scale. With $\ell_2$-normalized embeddings $\mathbf{v}^\top \mathbf{w}$ becomes the cosine similarity, and $\mathbf{p}_i(\mathbf{W})=(p_{i,c}(\mathbf{W}))_{1 \leq c \leq C}$ corresponds to softmax predicted scores vector. Contrastive VLMs allow zero-shot predictions, i.e., no need to learn $\mathbf{W}$, by embedding domain-specific textual descriptions for each label, e.g., "\texttt{An image of [cth class name]}.". Concretely, given a set of $J$ textual embeddings for each target category, $\{\{\mathbf{t}_{c,j}\}_{j=1}^{J}\}_{c=1}^{C}$, the zero-shot prototypes are the ensemble of the text embeddings for each class, $\mathbf{t}_{c}=\frac{1}{J}\sum_{j=1}^{J}\mathbf{t}_{c,j}$. These are integrated in Eq.~\eqref{eq:probs}, such that $\mathbf{W}^*=(\mathbf{t}_c)_{1 \leq c \leq C}$, which provides image predictions without requiring access to visual examples of the target task.

\subsection{Few-shot adaptation}
\label{ssec:fewshot}

Let us now denote a labeled set, $\mathcal{S}=\{i \in \mathbb{N} : 1 \leq i \leq N\}$, of visual labeled examples, $\{(\mathbf{v}_i, \mathbf{y}_i)\}_{i\in\mathcal{S}}$, a.k.a. \textit{support set}. Note that $\mathbf{y}$ are one-hot encoded vectors. In the few-shot literature, $N$ is typically small. Following standard notation, such a value is defined as proportional to the number of annotated examples ($K$) per category, a.k.a. shots, so that $N=K\times C$. However, it is worth noting that, as pointed out in recent works \cite{sstext25}, a realistic randomly retrieved support set should respect the intrinsic label-marginal distribution of the target data. 

Few-shot adaptation aims to use the support set to improve classification performance efficiently. Current solvers \cite{yu2023task,lin2023crossmodal,lp24,clap24,sstext25} involve multi-modal linear probes. These can be modeled from a constrained optimization perspective, where the learned prototypes, $\mathbf{W}$ in Eq.~\eqref{eq:probs}, are adjusted to minimize cross-entropy (CE) loss, but encouraged to remain close to the textual priors:
\begin{equation}
\label{eq:objective_fewshot_full}
    \phantom{.}\min_{\mathbf{W}} \
    \mathcal{L}(\mathbf{W}) = \frac{1}{|\mathcal{S}|} \sum\limits_{i\in \mathcal{S}} \mathcal{H}(\mathbf{y}_i,\mathbf{p}_i(\mathbf{W}))+ \sum\limits_{c\in\mathcal{Y}} \lambda^{\text{T}}_c ||\mathbf{w}_c - \mathbf{t}_c||_{2}^{2},
\end{equation}
where $\mathcal{H}(\mathbf{y}_i,\mathbf{p}_i(\mathbf{W}))=-\sum_{c\in\mathcal{Y}} y_{i,c} \ \text{ln}(p_{i,c}(\mathbf{W}))$ is the CE loss, and the second term is a class-weighted (by $\lambda^{\text{T}}_c>0$) $\ell_{2}$-penalty on the class weights deviation.

The standard softmax CE loss combines a tightness mechanism, $\mathcal{H}^{t}$, that approaches class prototypes toward their respective labeled embeddings, and a contrast term, $\mathcal{H}^{\text{cont}}$, that pushes them away from dense regions \cite{boudiaf2020unifying}:
\begin{equation}
\label{eq:entropy_parts}
\mathcal{H}(\mathbf{y}_i,\mathbf{p}_i(\mathbf{W})) = \underbrace{ - \sum_{c\in \mathcal{Y}} y_{i,c} \ (\mathbf{v}_i^\top \mathbf{w}_{c} / \tau)}_{\mathcal{H}^{t}(\mathbf{y}_i,\mathbf{W})} + \underbrace{\sum_{c\in \mathcal{Y}} y_{i,c} \ \text{ln} ( \sum_{j\in \mathcal{Y}} \text{exp}(\mathbf{v}_i^\top \mathbf{w}_{j} / \tau) )}_{\mathcal{H}^{\text{cont}}(\mathbf{W})}.
\end{equation}
Recent literature suggests that the contrastive term might degrade generalization of linear probes, especially in few-shot imbalanced scenarios \cite{sstext25}. Indeed, the idea of relying solely on a hard-label assignment in low-shot regimes builds upon early observations on the robustness of average class prototypes for few-shot classification \cite{simpleshot}, a.k.a. Simple-Shot (SS). Considering only the tightness term, the few-shot learning objective for VLMs in Eq.~\eqref{eq:objective_fewshot_full} reads:
\begin{equation}
\label{eq:objective_fewshot}
    \phantom{.}\min_{\mathbf{W}} \
    \mathcal{L}_{\text{FEW-SHOT}}(\mathbf{W}) = \frac{1}{|\mathcal{S}|} \sum\limits_{i\in \mathcal{S}} \mathcal{H}^{t}(\mathbf{y}_i,\mathbf{W}) + \sum\limits_{c\in\mathcal{Y}} \lambda^{\text{T}}_c ||\mathbf{w}_c - \mathbf{t}_c||_{2}^{2}.
\end{equation}
As first proposed in SS-Text \cite{fca25,sstext25}, the objective in Eq. \ref{eq:objective_fewshot} has a closed-form solution, which is attractive for validation-free settings compared to hyper-parameter-based gradient descent solvers. 

\section{Improving low-shot adaptation with unlabeled data}
\label{sec:our_method}

Let us denote an extended adaptation set, $\mathcal{D}=\{i \in \mathbb{N} : 1 \leq i \leq N+M\}=\mathcal{S} \cup \mathcal{U}$, where $\ \mathcal{U}:|\mathcal{U}|=M$ is an unlabeled set, in which only access to the visual embeddings is granted. This extension of the vanilla few-shot scenario opens the door to incorporating semi-supervised learning \cite{chapelle2009semi}, e.g., via pseudo-labels \cite{lee2013pseudo}, in our case generated by label propagation from zero-shot textual priors in low-shot regimes. In the following, we describe our novel solver, \textbf{SS-Text-U}, which efficiently integrates both labeled, textual, and unlabeled supervisory signals.

\subsection{Proposed objective function}
\label{ssec:objective_semisup}

\noindent{\bfseries Learning from unlabeled data.} Let us denote $\mathbf{z}_i = (z_{i,k})_{(1 \leq c \leq C)} \in \triangle_C , i\in\mathcal{U}$ as the sample-to-class assignment variables within the probability simplex in the unlabeled data. We aim to use such pseudo-labels as supervisory signals during adaptation. In particular, our constrained unsupervised loss aims to minimize the tightness-CE loss between predictions and pseudo-labels, while enforcing its label distribution to present a consistent structure:
\begin{equation}
\label{eq:objective_unlabeled}
    \mathcal{L}_{\text{U}}(\mathbf{W},\mathbf{z}) = \frac{1}{|\mathcal{U}|} \sum\limits_{i\in \mathcal{U}} \mathcal{H}^{t}(\mathbf{z}_i,\mathbf{W}) \ \text{s.t.} \ \hat{\mathbf{m}}=\mathbf{m},
\end{equation}
where $\mathbf{m}=\sum_{i\in \mathcal{S}}\mathbf{y}_i / N$ is the true label distribution estimated from the support set, and $\hat{\mathbf{m}}=\sum_{i\in \mathcal{U}}\mathbf{z}_i / \sum_{i\in \mathcal{U}, c\in \mathcal{Y}}z_{i,c}$ is the one obtained from the $\mathbf{z}$ assignments.

\noindent{\bfseries Combined objective.} Our objective loss is the weighted sum of the few-shot criteria in Eq.~\eqref{eq:objective_fewshot} and the unlabeled term in Eq.~\eqref{eq:objective_unlabeled}, and it considers two parameters: the class prototypes and the label assignments on the unlabeled data:
\begin{equation}
\label{eq:objective_semisup}
    \phantom{.}\min_{\mathbf{W},\mathbf{z}} \
    \mathcal{L}_{\text{SEMI}}(\mathbf{W},\mathbf{z}) = \mathcal{L}_{\text{FEW-SHOT}}(\mathbf{W}) + \mathbf{\lambda^{\text{U}}} \mathcal{L}_{\text{U}}(\mathbf{W},\mathbf{z}).
\end{equation}

\subsection{Block-wise optimization}
\label{ssec:optimizer}

As our objective depends on two variables $(\mathbf{W},\mathbf{z})$, we adopt an inexact block coordinate minimization (BCM) \cite{tseng2001convergence,razaviyayn2013unified,wright2015coordinate}. At iteration $t=1,\dots,T$, the solver alternates updating $\mathbf{z}$ (approximately solving its subproblem) and computing a closed-form update for $\mathbf{W}$, each while keeping the other block fixed. Each step (approximately) decreases the objective, resulting in monotonic optimization.

\noindent{\bfseries \circled{1}~Z-block updates.} For a fixed prototypes matrix, $\mathbf{W}^{t-1}$, the optimization of $\mathbf{z}$ only depends on the unlabeled loss term in our combined objective in Eq.~\eqref{eq:objective_semisup}:
\begin{equation}
\label{eq:objective_unlabeled_extended}
    \phantom{.}\min_{\mathbf{z}} \ \mathcal{L}_{\text{U}}(\mathbf{W}^{t-1},\mathbf{z}) = - \frac{1}{|\mathcal{U}|} \sum\limits_{i\in \mathcal{U}} \sum\limits_{c\in \mathcal{Y}} z_{i,c} \ (\mathbf{v}_i^\top \mathbf{w}_{c}^{t-1}  / \tau) \ \text{s.t.} \ \hat{\mathbf{m}}=\mathbf{m}.
\end{equation}
Let $\mathbf{S}\in \mathbb{R}^{C\times|\mathcal{U}|}$ be the similarity matrix of the unlabeled set, with entries $\mathbf{S}=((\mathbf{v}_i^\top \mathbf{w}_{c}^{t-1} / \tau)_{c \in \mathcal{Y}})_{i \in \mathcal{U}}$, and $\mathbf{Q}=(\mathbf{z}_i)_{i \in \mathcal{U}}$, $\mathbf{Q}\in \mathbb{R}^{C\times|\mathcal{U}|}$ denote the pseudo-label assignments to be learned. We formulate the optimization of Eq.~\eqref{eq:objective_unlabeled_extended} as a constrained similarity-maximization problem, i.e., $\max_{\mathbf{Q} \in \mathcal{Q}} \; tr(\mathbf{Q}^\top \mathbf{S})$, $\text{s.t.} \ \hat{\mathbf{m}}=\mathbf{m}$. This can be cast as an \textit{Optimal Transport} problem \cite{ot_book,cuturi2013sinkhorn}, where $\mathbf{Q}$ is relaxed to be an element of the transportation polytope $\mathcal{Q}=\{ \mathbf{Q} \mid \mathbf{Q} \mathbf{1}_{(|\mathcal{U}|)} = \mathbf{m}, \mathbf{Q}^\top \mathbf{1}_K = \mathbf{u}_{(|\mathcal{U}|)} \}$,
where $\mathbf{1}_{(\cdot)}$ denotes a column vector of ones, and $\mathbf{u}_{(\cdot)}$ an uniform distribution, being $(\cdot)$ the length. To solve this efficiently, we resort to the Sinkhorn-Knopp algorithm \cite{cuturi2013sinkhorn}, which enforces an entropy-regularized structure on the transport plan, yielding the assignment:
\begin{equation}
\label{eq:codes}
\mathbf{Q}^* = \text{Diag}(\mathbf{r}^{(T_{\text{OT}})}) \mathbf{Q}^{(0)} \text{Diag}(\mathbf{c}^{(T_{\text{OT}})}).
\end{equation}
The renormalization vectors are computed using a small number of matrix multiplication iterations, $t_{\text{OT}}=1, ..., T_{\text{OT}}$, where in each iteration:
\begin{align}
   & \mathbf{r}^{(t_{\text{OT}})}=\mathbf{m}/(\mathbf{Q}^{(0)} \mathbf{c}^{(t_{\text{OT}}-1)}),
   & \mathbf{c}^{(t_{\text{OT}})}=\mathbf{u}_{(|\mathcal{U}|)}/(\mathbf{Q}^{(0)} \mathbf{r}^{(t_{\text{OT}})}),   \label{renorm_2} 
\end{align}
with $\mathbf{c}^{(0)}=\mathbf{1}_{(|\mathcal{U}|)}$, and $\mathbf{Q}^{(0)}=\exp(\mathbf{S})/\sum(\exp(\mathbf{S})$. Upon convergence, the normalized soft codes, $\mathbf{z}_i^{*}$ (columns in $\mathbf{Q}^{*}$), are the block-updated results, $\mathbf{z}^{t}$.

\noindent{\bfseries \circled{2}~Closed-form updates for $\mathbf{W}$.} When $\mathbf{z}$ is fixed in a given BCM iteration, i.e., $\mathbf{z}=\mathbf{z}^{t}$, our objective in Eq.~\eqref{eq:objective_semisup} is convex. It can be minimized by setting its gradient w.r.t. each $\mathbf{w}_{c}$ to zero, which yields the following closed-form update:
\begin{equation}
\label{eq:solver_wc_semisup}
\mathbf{w}_{c}^{t} = \arg\min_{\mathbf{w}_c} \ \frac{\partial \mathcal{L}_{\text{{SEMI}}}(\mathbf{W},\mathbf{z}^{t})}{\partial \mathbf{w}_c} = \frac{1}{2 \lambda^{\text{T}}_c \ |\mathcal{S}| \ \tau}\sum\limits_{i\in \mathcal{S}} y_{i,c} \mathbf{v}_i + \frac{\lambda^{\text{U}}_c}{2 \lambda^{\text{T}}_c \ |\mathcal{U}| \ \tau}\sum\limits_{i\in \mathcal{U}} z_{i,c}^{t} \mathbf{v}_i + \mathbf{t}_c.
\end{equation}

\noindent{\bfseries Implementation.} \textbf{\circled{1}}~For the BCM initialization, we use $\mathbf{w}_c^{0}=\mathbf{t}_c$. The number of iterations is set to $T=3$. \textbf{\circled{2}}~Regarding the $\mathbf{z}$-block updates (Eq.~\eqref{eq:codes}), we set $T_{\text{OT}}=10$. Another element to consider in the z-block updates is the label-marginal distribution, $\mathbf{m}$, estimated from the support set. In imbalanced low-shot regimes ($K\in\{1, 2\}$), we might obtain non-represented categories, i.e., $m_c=0$. Hence, we use a post-processing, where a baseline, $b$, is provided to all labels, i.e., $\bar{m}_c=max(m_c,b)$, $b$ fixed to be smaller than the least-observed frequent category, i.e., $b=r\cdot \min_{c:m_c!=0}(\mathbf{m}), 0<r<1$. We set its default value to $r=1/4$. \textbf{\circled{3}}~For the $\mathbf{W}$ update (Eq.~\eqref{eq:solver_wc_semisup}), we fix $\lambda^{\text{T}}_c=(1/K_c)=(1/\sum_{i\in \mathcal{S}}y_{i,c}) \ ; \ \lambda^{\text{U}}_c=2\lambda^{\text{T}}_c$. The goal is to provide a data-driven design that increases the importance of the support set in high-shot regimes while relying on textual prototypes when $\sum_{i\in \mathcal{S}}y_{i,c}\rightarrow0$. Also, our $\lambda^{\text{U}}_c$ aims to simplify the solver and ensure the significance of the unlabeled data in low-shot regimes.

\section{Experiments}
\label{sec:experiments}

\subsection{Setup}
\label{ssec:setup}

\begin{figure*}[t!]
    \begin{center}
        \begin{tabular}{cccc}

         \includegraphics[width=.24\linewidth]{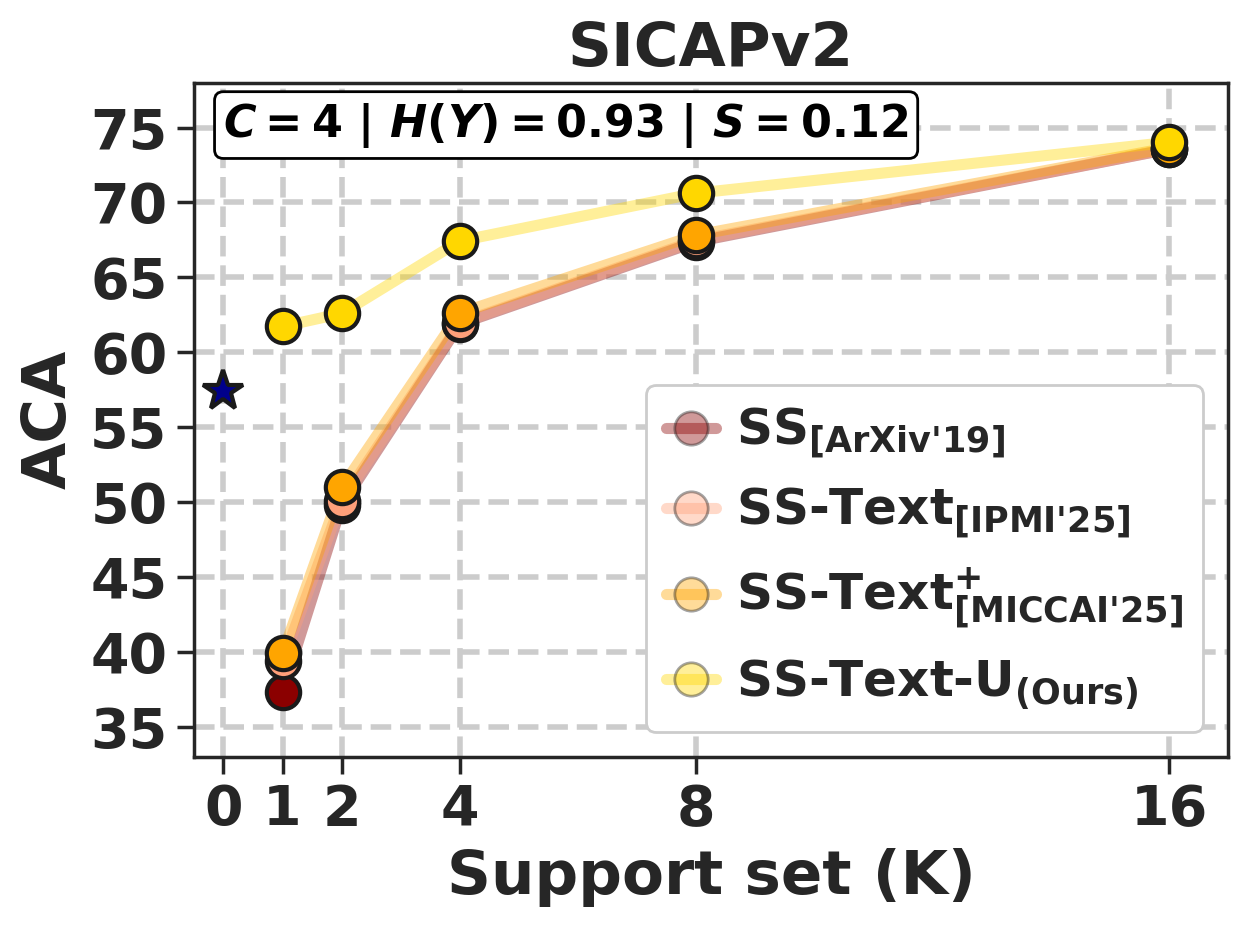}       &
         \includegraphics[width=.24\linewidth]{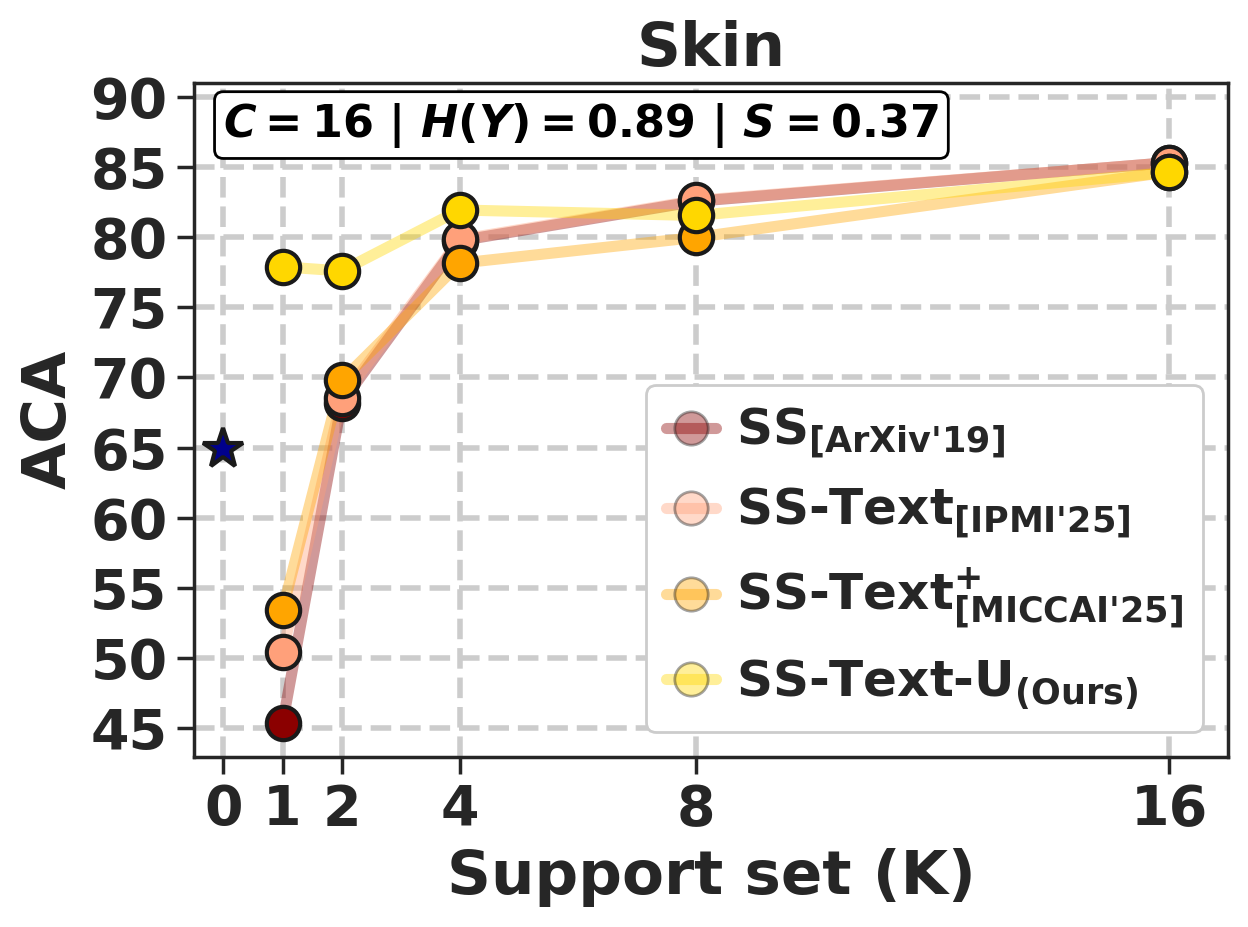}          &
         \includegraphics[width=.24\linewidth]{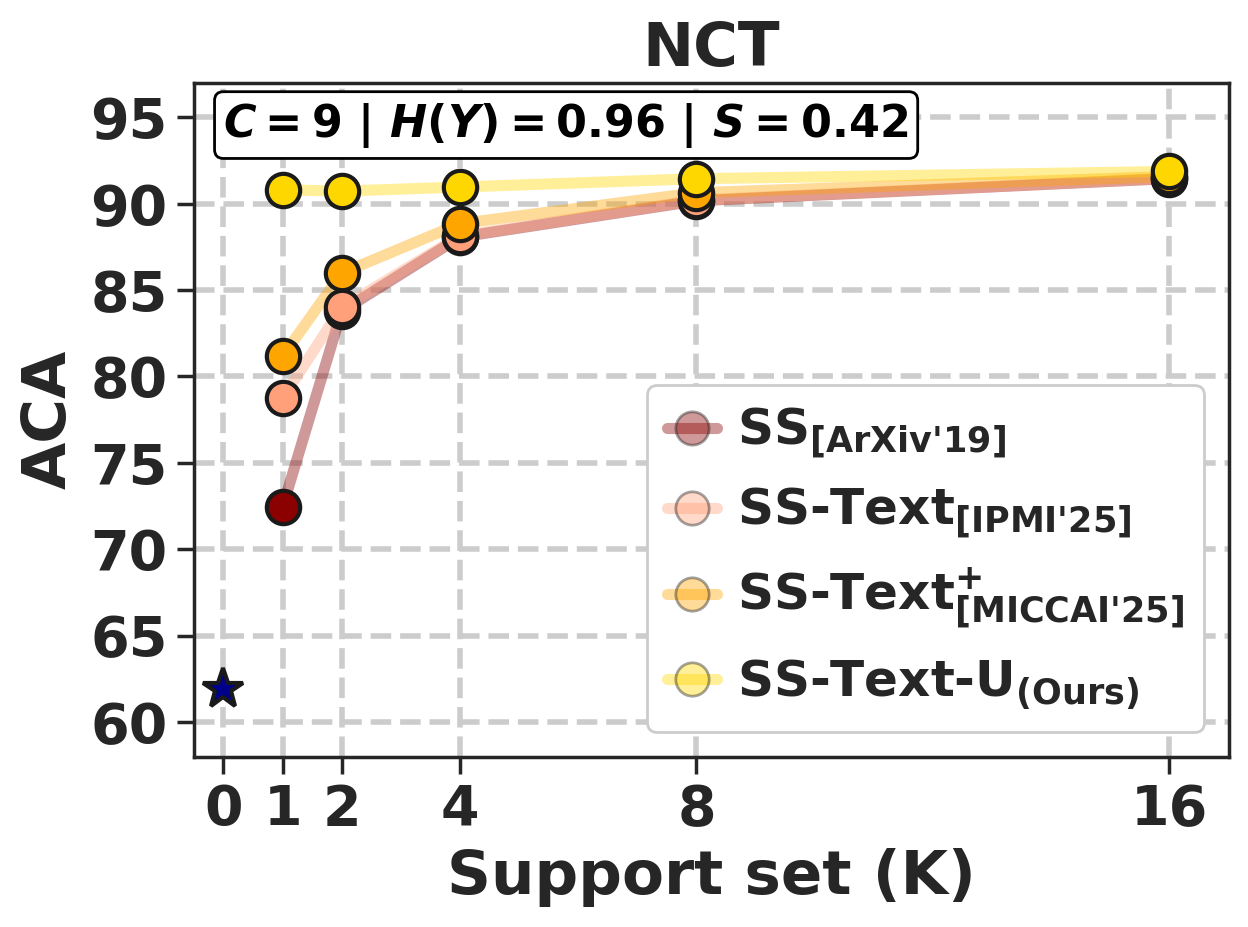}           & 
         \includegraphics[width=.24\linewidth]{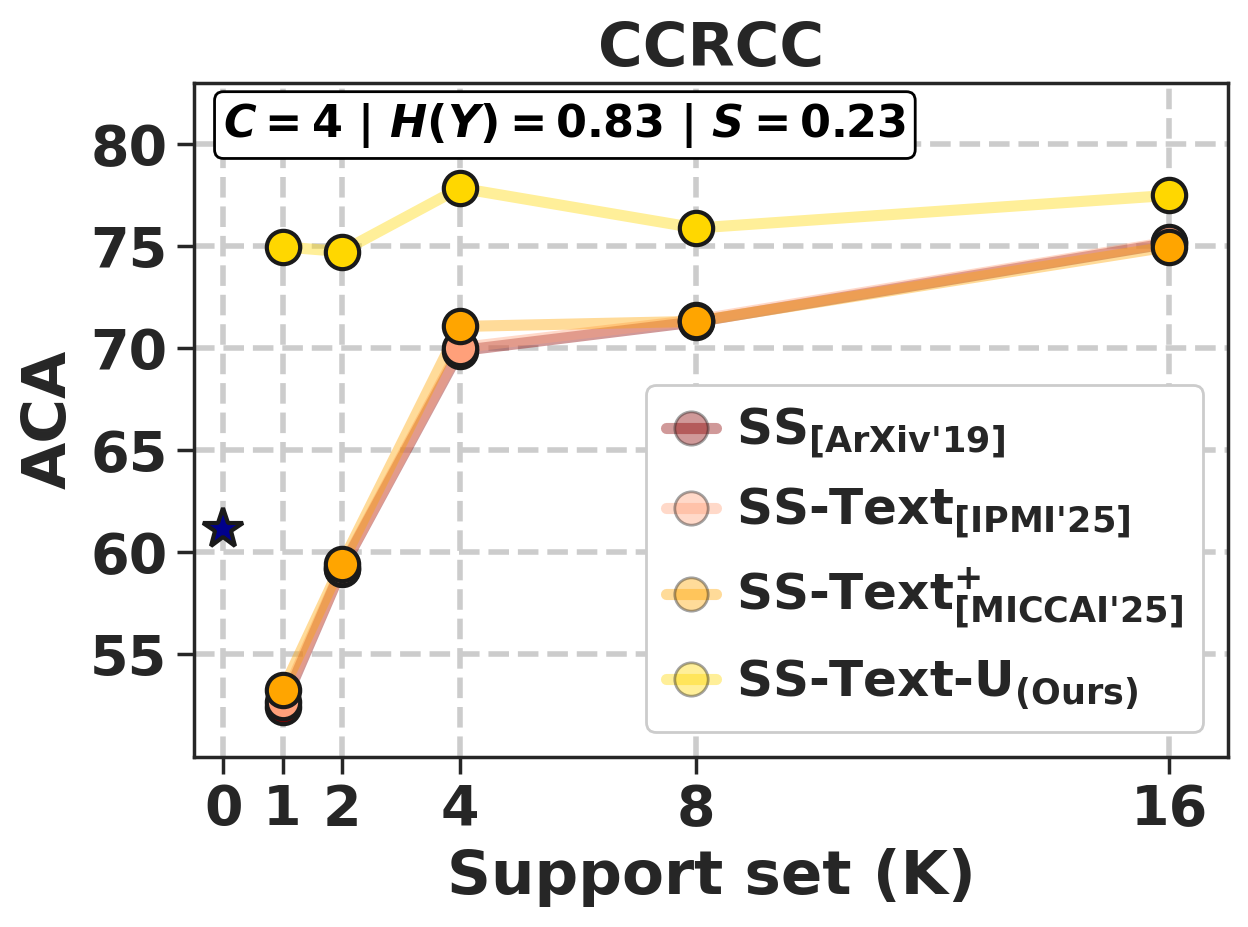}           \\

         \includegraphics[width=.24\linewidth]{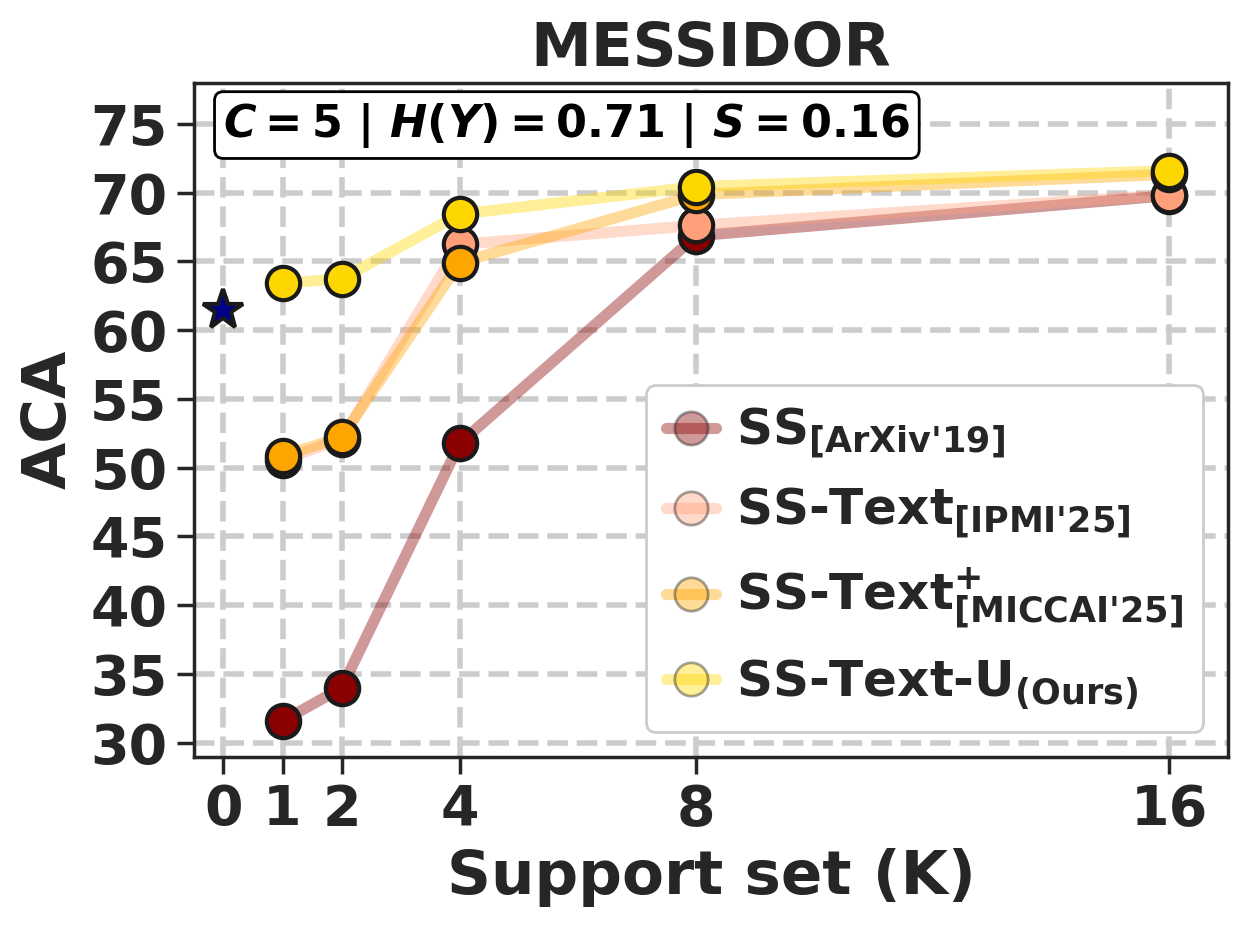}      &
         \includegraphics[width=.24\linewidth]{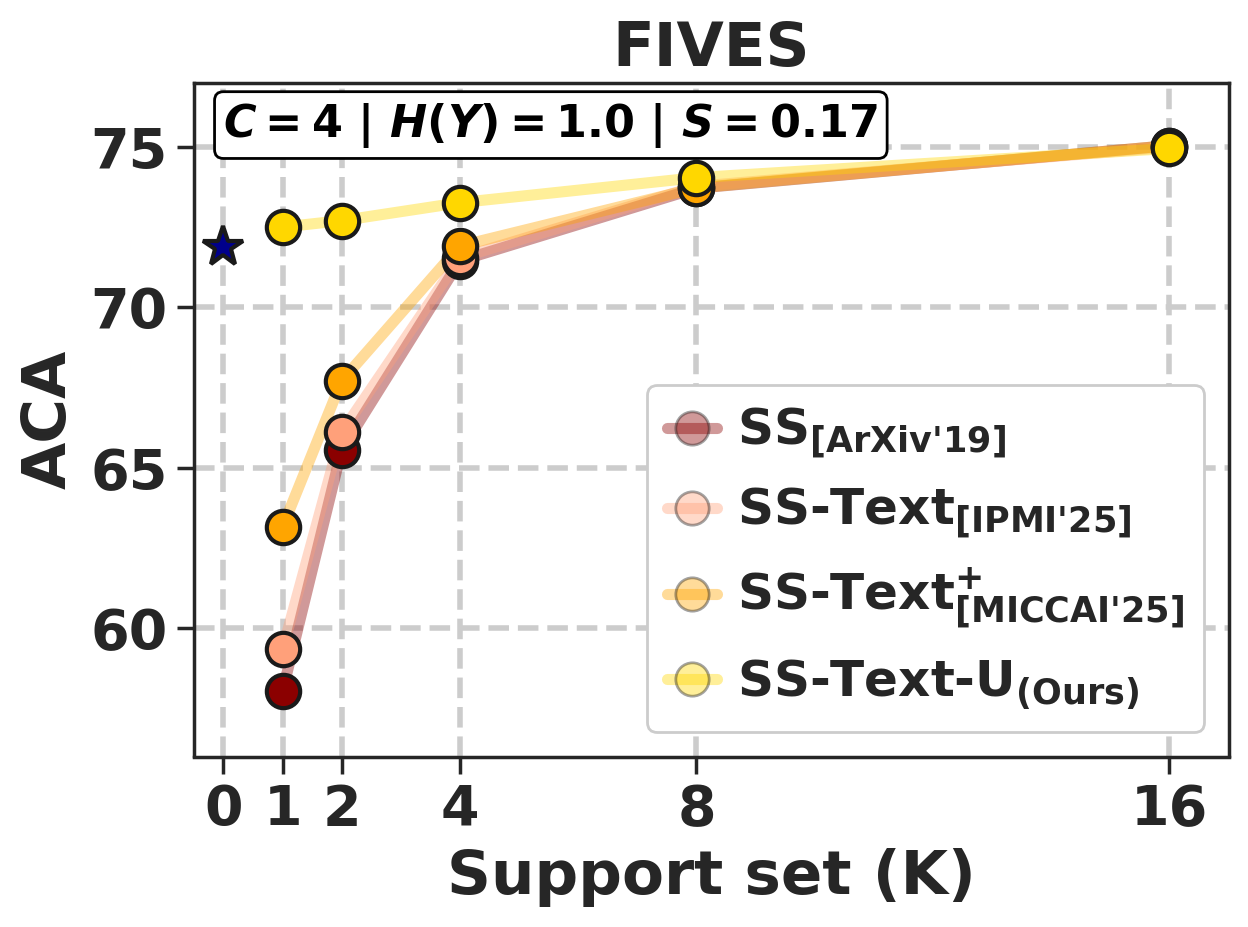}         &
         \includegraphics[width=.24\linewidth]{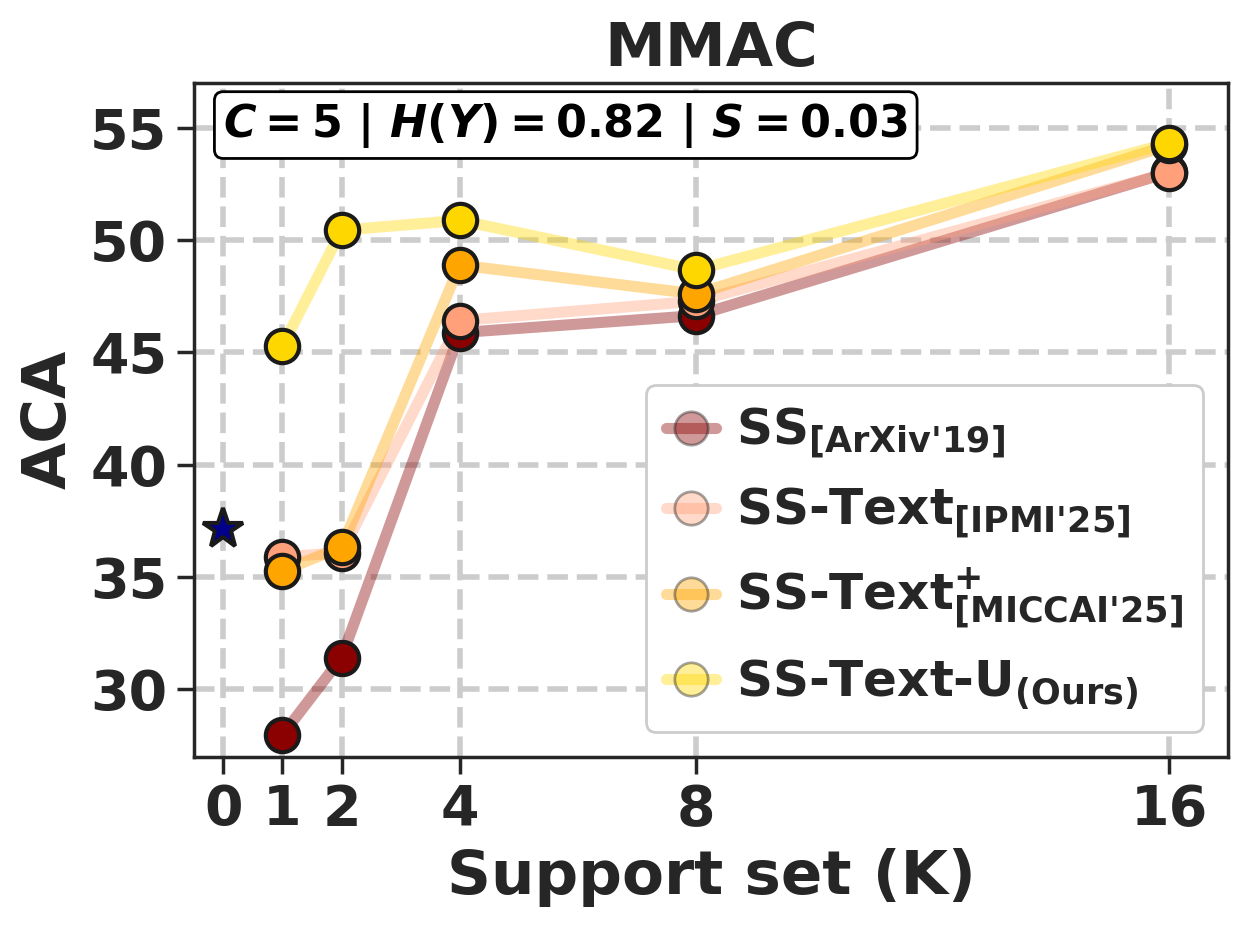}          &
         \includegraphics[width=.24\linewidth]{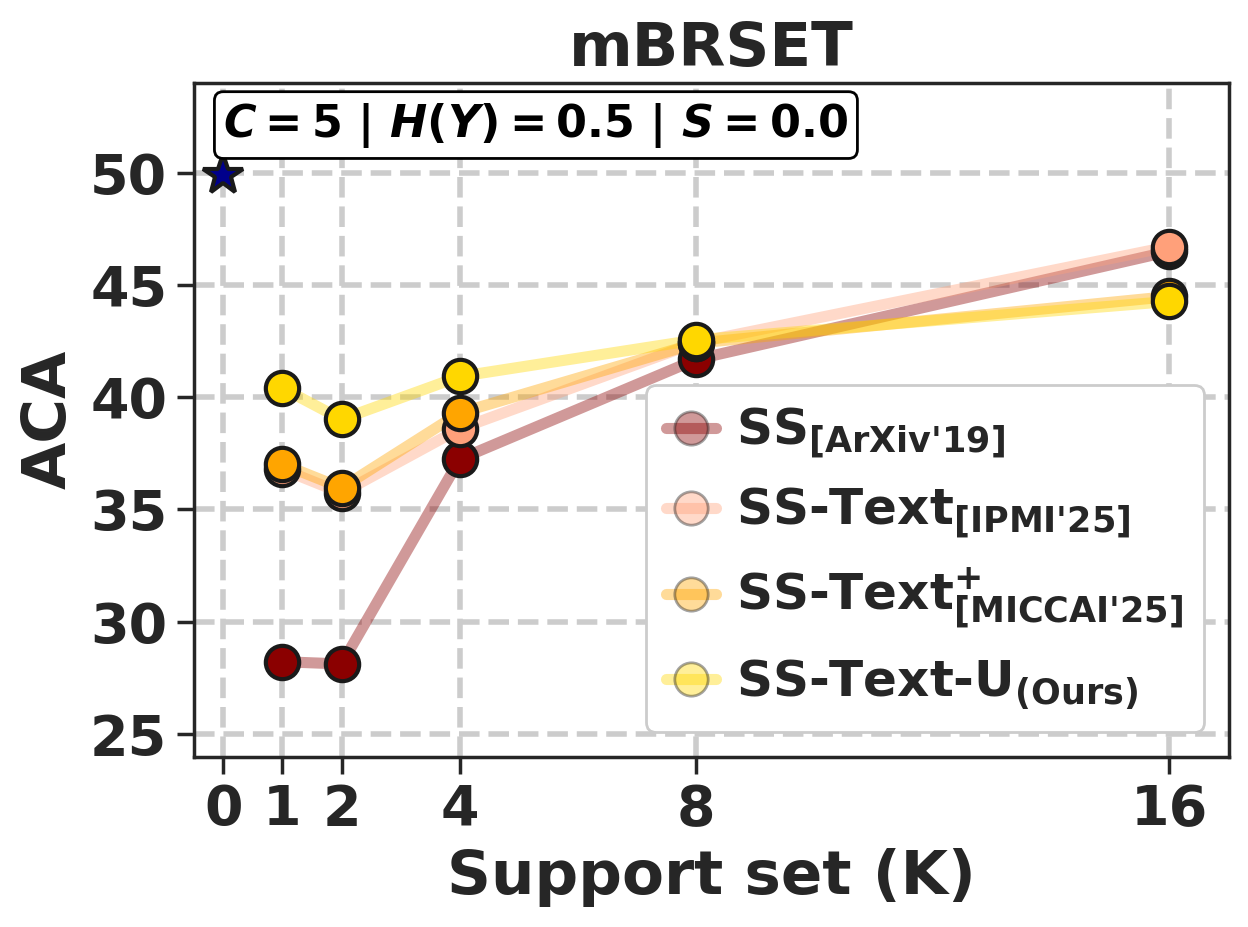}          \\

         \includegraphics[width=.24\linewidth]{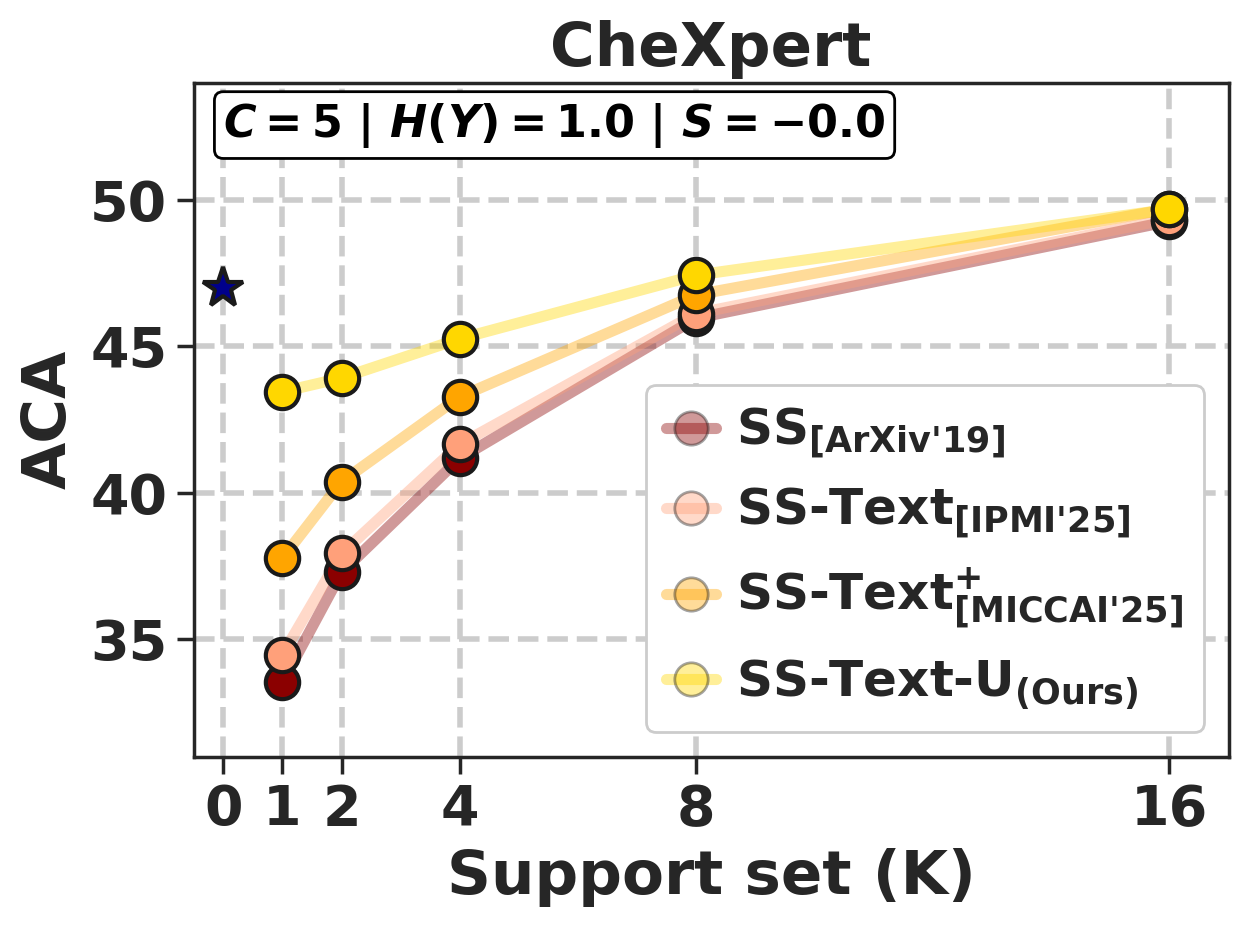} &
         \includegraphics[width=.24\linewidth]{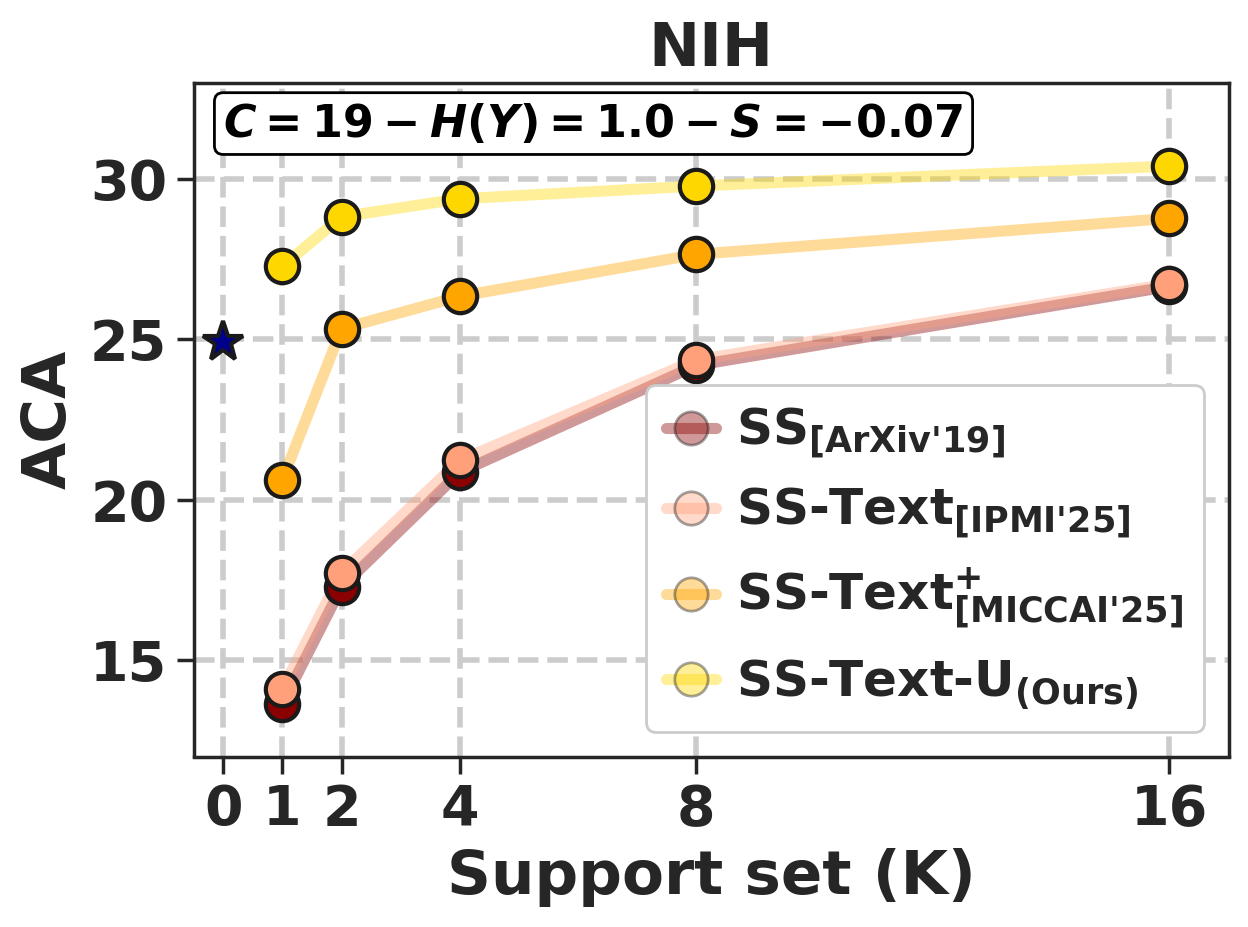}           &
         \includegraphics[width=.24\linewidth]{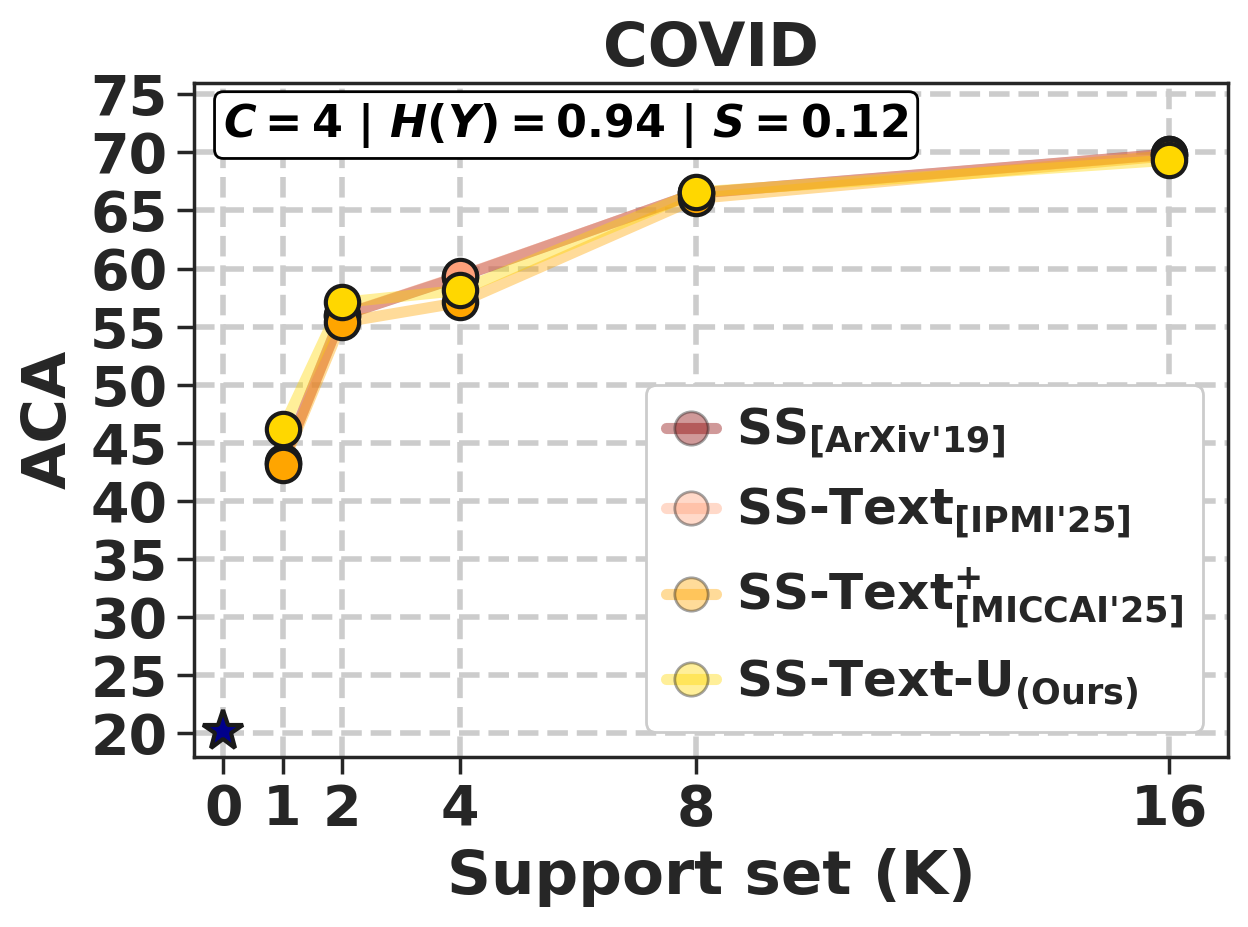}         &
         \includegraphics[width=.24\linewidth]{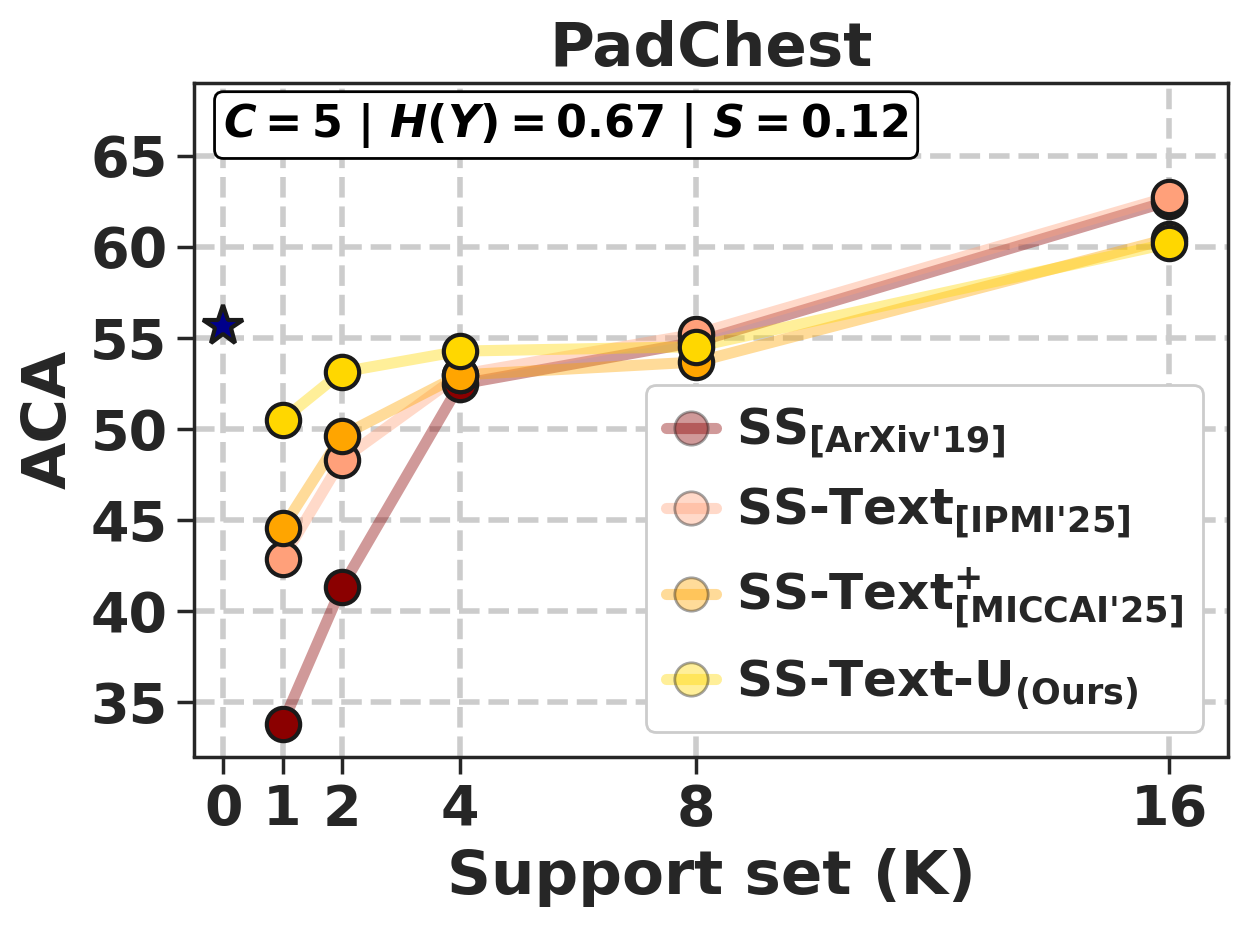}          \\

        \end{tabular}
        \caption{\textbf{Few-shot adaptation performance per dataset}.}
        \label{fig:per_dataset}
    \end{center}
\end{figure*}

\noindent\textbf{{Modalities, VLMs and transfer tasks}.} We follow up on \cite{sstext25}: three medical imaging modalities are tackled, for which a pre-trained VLM is adapted to several downstream tasks. We extend the benchmark presented in \cite{sstext25}, and incorporate one additional task per modality. For \textbf{histology}, we use CONCH~\cite{CONCH}, transferred to tissue classification and tumor detection/grading in NCT-CRC~\cite{kather2018100}, SICAPv2~\cite{silva2020going}, SkinCancer~\cite{kriegsmann2022deep}, and CCRCC~\cite{CCRCC}. We employ FLAIR~\cite{FLAIR} as the \textbf{ophthalmology} foundation model for color fundus images, adapted to disease detection/grading tasks in MESSIDOR~\cite{messidor}, MMAC~\cite{mmac}, FIVES~\cite{fives}, and mBRSET~\cite{mBRSET}. Finally, for \textbf{radiology} (chest X-ray), we use CONVIRT~\cite{convirt} with weights from \cite{sstext25}, adapted for disease/findings multi-class classification on CheXpert~\cite{irvin2019chexpert}, NIH-LT~\cite{nih,nih_lt}, COVID~\cite{covid1,covid2}, and PadChest~\cite{Bustos2019PadChest:Reports}. The number of classes per dataset (C) and their imbalanced ratio (in terms of normalized label-marginal entropy, $\mathcal{H}(Y)\in(0, 1)$ - max indicating a balanced distribution) are presented in Fig~\ref{fig:per_dataset}. We used the text templates proposed by the authors of each VLM in their respective papers/repositories. For specific details on data splits, text prompts, and class names, we refer to our project's repository: \href{https://github.com/jusiro/SS-Text-U}{SS-Text-U}.

\noindent\textbf{{Baselines}.} Relevant few-shot solvers are explored. First, gradient-based linear probes are used: vanilla ZS-LP~\cite{clap24}, and SoTA text-constrained solutions such as CLAP~\cite{clap24} and LP++~\cite{lp24,shakeri2024few}. Second, we evaluate so-called training-free solutions: Simple-shot (SS)~\cite{simpleshot}, SS-Text~\cite{fca25}, and SS-Text+~\cite{sstext25}. For all methods, we stick to the validation-free scenario from \cite{sstext25}, and keep the training hyper-parameter fixed across datasets, using the learning schedules detailed in \cite{sstext25}.

\noindent\textbf{{Evaluation protocol}.} We sample imbalanced, realistic labeled support sets as in \cite{sstext25} with $K\in\{1, 2, 4, 8, 16\}$, as standard in the literature \cite{clap24,shakeri2024few,sstext25}. Regarding the unlabeled data pool, $M=C\times24$ additional data points are sampled. Class-wise balanced accuracy (ACA) is employed as a figure of merit, as recommended in \cite{metrics}. All results are the average across 50 different random seeds for sampling.

\subsection{Results}
\label{ssec:results}

\noindent\textbf{{Main results}.} Figure~\ref{fig:motivation}(b) shows the benefit of leveraging unlabeled data via the proposed SS-Text-U solver, which consistently brings average performance gains over all few-shot baselines. Compared to the best training-free solver (SS-Text+), the gains are significant: $10.9\%$, $7.1\%$, $2.7\%$, $1.3\%$, and $0.3\%$ in ACA across increasing shots. From a practical standpoint, these numbers translate into a decrease in the required labeling effort of nearly $50\%-75\%$ for $K\in\{2, 4, 8\}$, e.g., SS-Text-U using 1-shot performs similarly to SS-Text+ when $K=4$.

\begin{figure*}[t!]
    \begin{center}
        \begin{tabular}{ccc}

        \multicolumn{2}{c}{
        \begin{tabular}{cc}
        \includegraphics[width=.23\linewidth]{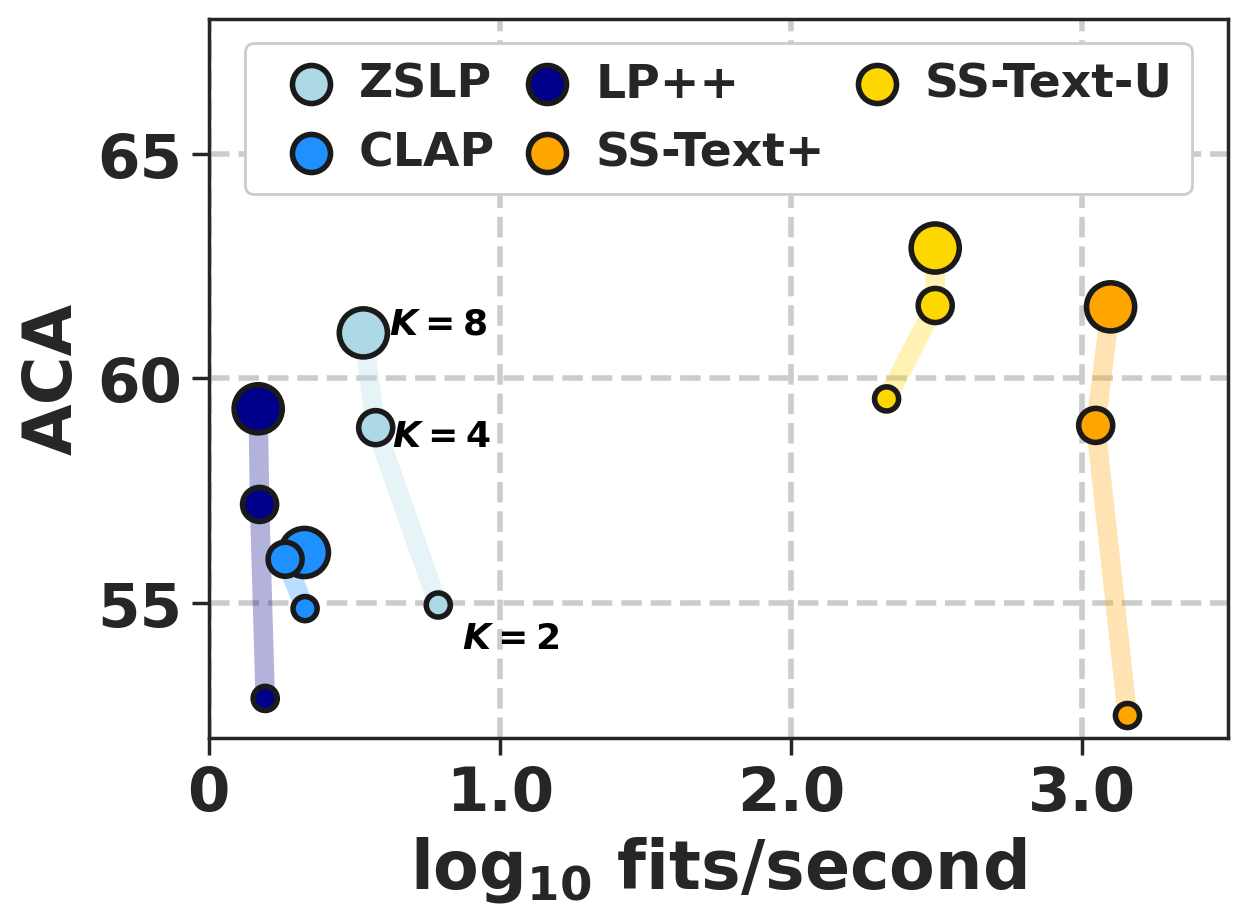} &
        \includegraphics[width=.23\linewidth]{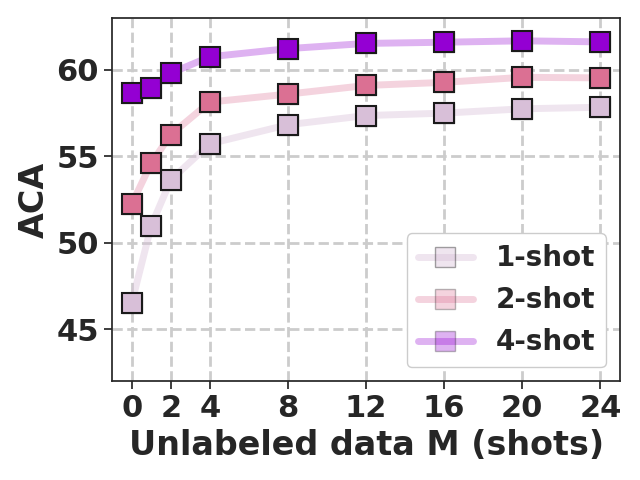} \\
        \scriptsize{\textbf{(a) Speed}} & \scriptsize{\textbf{(b) Data-efficiency}} \\
        \includegraphics[width=.23\linewidth]{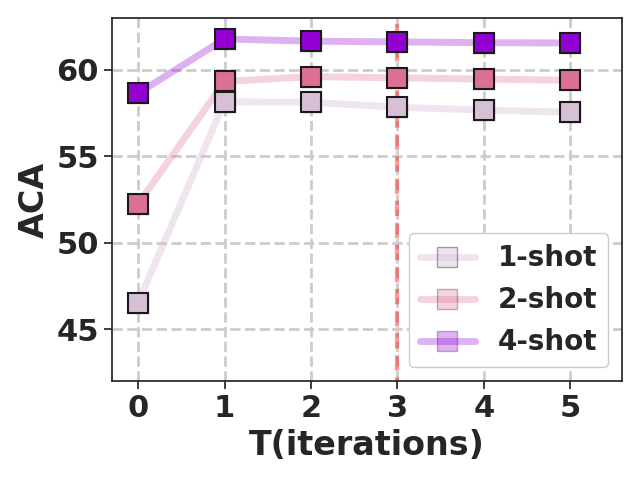} &
        \includegraphics[width=.23\linewidth]{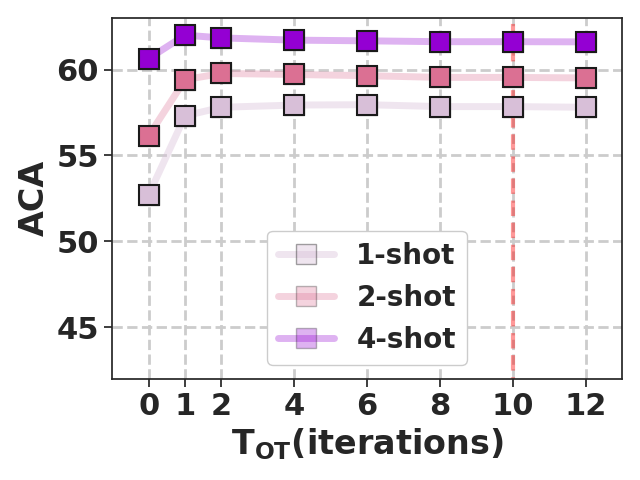} \\
        \multicolumn{2}{c}{\scriptsize{\textbf{(c) Convergence}}}
        \end{tabular}
        } &
        
        \multicolumn{1}{c}{
        \begin{tabular}{c}
        \includegraphics[width=.35\linewidth]{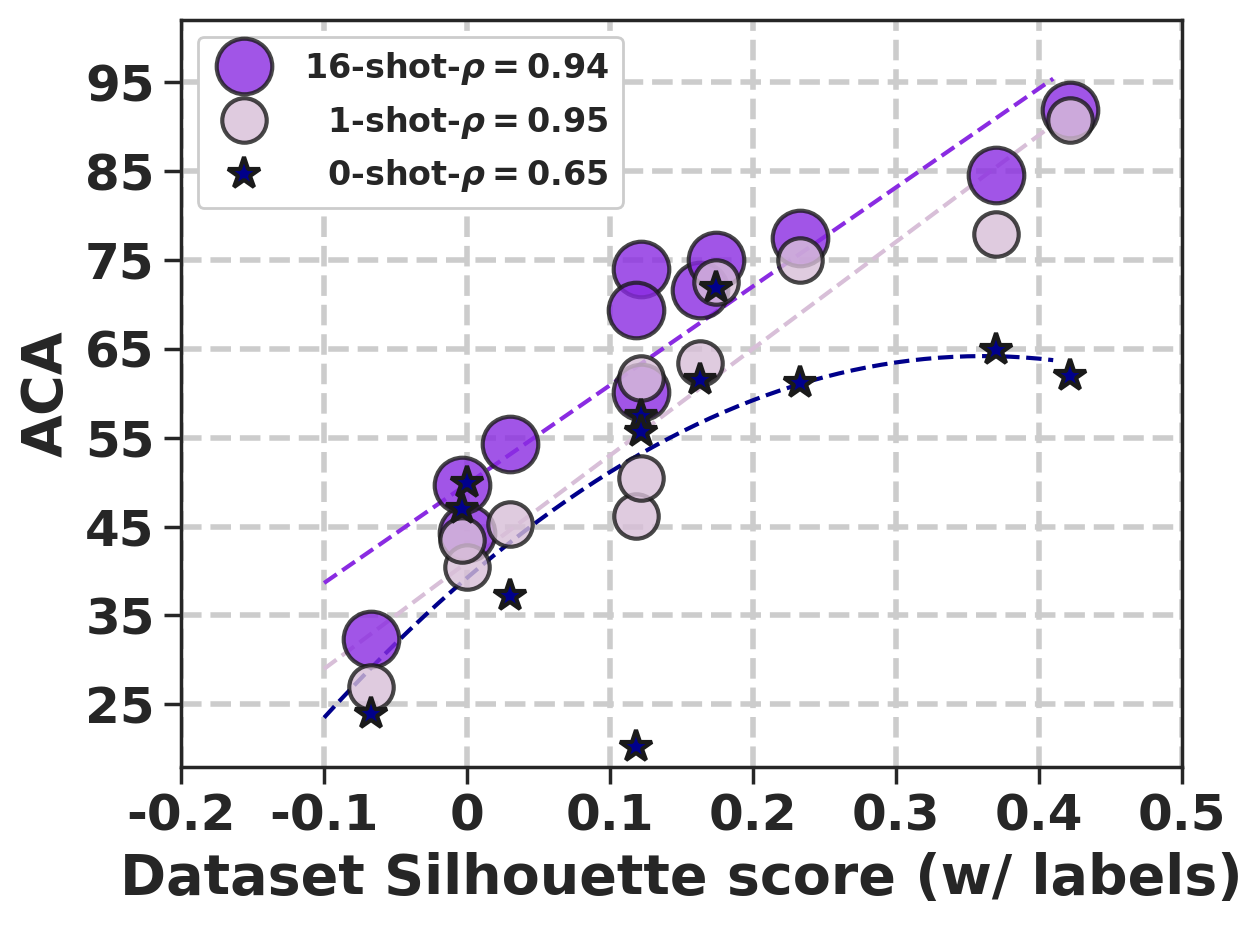} \\
        \scriptsize{\textbf{(d) Exploratory analysis}}                          
        \end{tabular}
        } \\
 
        \end{tabular}
        \caption{\textbf{Studies on efficiency, convergence, and exploratory analysis}.}
        \label{fig:ablation_studies}
    \end{center}
\end{figure*}

\noindent\textbf{{Efficiency}.} As shown in Figure~\ref{fig:ablation_studies}(a), the proposed solver does not incur significant computation overhead, being orders of magnitude faster than gradient-based few-shot approaches. For example, for large datasets, e.g., Skin, it takes approximately $25$ \si{\milli\second} using a commodity laptop without specialized software.

\noindent\textbf{{How much unlabeled data is necessary}?} We answer this question in Figure~\ref{fig:ablation_studies}(b). Interestingly, using only $M=C\times8$ shots of unlabeled samples already brings a noteworthy improvement in low-shot regimes.

\noindent\textbf{{Importance of our constraint pseudo-labels}}. Figure~\ref{fig:ablation_studies}(c) (\textit{right}) shows the importance of considering the structure of the label distribution in the pseudo-labels as proposed in Eq~\ref{eq:objective_unlabeled}, by comparing executing $0$ iterations of the Sinkhorn Optimal Transport against $T_{\text{OT}}\geq1$. For example, for $K=1$ and $K=4$, average performance gains of nearly $5.1\%$ and $1.0\%$ are obtained, respectively.

\noindent\textbf{{Convergence}.} Also, in Figure~\ref{fig:ablation_studies}(c), one can notice the stability in convergence of the BCM solver (\textit{left}) and its Optimal Transport stage during z-block updates (\textit{right}), as well as the iterations used as default for $T$ and $T_{\text{OT}}$ (\textit{dashed red lines}), which are set high enough to ensure convergence of both iterative processes.

\noindent\textbf{{Hyper-parameters}.} Table~\ref{ablation_studies}(a) motivates the design choice of $\lambda^{\text{T}}_c$ and $\lambda^{\text{U}}_c$. First, it shows (\textit{top}) the motivation for our adaptive design choice to properly scale across data regimes compared to using fixed values of $\lambda^{\text{T}}_c$, which are more rigid in performance, e.g., high values do not perform well on larger-shot regimes, and vice-versa. Second, it shows (\textit{bottom}) that using fixed values of $\lambda^{\text{U}}_c$ produces significant performance drops. Table~\ref{ablation_studies}(b) shows the effect of the correction ratio, $r$, to correct the target label distribution, $\mathbf{m}$, for missing categories when $K\in\{1, 2\}$, as detailed in Section~\ref {ssec:objective_semisup}. One can readily notice that, in terms of ACA, different values, e.g., $r\in\{1/2, 1/4, 1/8\}$, provide performance close to the oracle scenario (known $\mathbf{m}$), and \textit{all improve over the SS-Text+ baseline}. We set the default value to $r=1/4$, as smaller values (which enforce an overly large appearance of under-represented classes) may penalize the overall performance when imbalanced-sensitive metrics, such as accuracy (Acc), are considered.

\begin{table}[t!]
\setlength{\tabcolsep}{1.8pt}
\centering
{\fontsize{8}{9}\selectfont
\caption{\textbf{Ablation studies on SS-Text-U hyper-parameters.}}
\label{ablation_studies}
\begin{tabular}{ cc }
    \begin{tabular}{c}
        \begin{tabular}{lcccc}
            \toprule
            \multicolumn{1}{l}{\multirow{1}{*}{Setting $\downarrow$  /  $K$ $\rightarrow$}} & $1$     & $2$    & $8$    & $16$   \\
            \midrule
            $\lambda^{\text{T}}_c=0.1 \ ; \ \lambda^{\text{U}}_c=2\lambda^{\text{T}}_c$                      & | 52.6 & 56.3 & 63.0 & 65.9   \\
            $\lambda^{\text{T}}_c=1.0 \ ; \ \lambda^{\text{U}}_c=2\lambda^{\text{T}}_c$                      & | 58.0 & 60.2 & 62.3 & 63.1   \\
            $\lambda^{\text{T}}_c=10 \ \ ; \ \lambda^{\text{U}}_c=2\lambda^{\text{T}}_c$                     & | 57.4 & 57.6 & 56.9 & 56.8   \\
            \rowcolor{Gray} $\lambda^{\text{T}}_c=1/K_c \ ; \ \lambda^{\text{U}}_c=2\lambda^{\text{T}}_c$    & | 57.8 & 59.5 & 62.9 & 65.4   \\
            \midrule
            $\lambda^{\text{T}}_c=1/K_c \ ; \ \lambda^{\text{U}}_c=0.1$                                      & | 47.5 & 52.8 & 60.2 & 64.1   \\
            $\lambda^{\text{T}}_c=1/K_c \ ; \ \lambda^{\text{U}}_c=1.0$                                      & | 49.3 & 55.4 & 62.2 & 65.1   \\
            $\lambda^{\text{T}}_c=1/K_c \ ; \ \lambda^{\text{U}}_c=10$                                       & | 50.1 & 55.5 & 60.7 & 60.3   \\
            \midrule
        \end{tabular} \\
    \textbf{(a) Study on $\lambda^{\text{T}}_c$ and $\lambda^{\text{U}}_c$ in Eq.~\eqref{eq:solver_wc_semisup}.}
    \end{tabular}
    &
    \begin{tabular}{c}
        \begin{tabular}{lccc}
            \toprule
            \multicolumn{1}{l}{\multirow{1}{*}{$K$ $\rightarrow$}} & $1$     & $2$    & $4$ \\
            \midrule
            $r=1/2$                   & | 58.5/62.5 & 60.1/65.6 & 61.7/68.7 \\
            \rowcolor{Gray} $r=1/4$   & | 57.8/65.5 & 59.5/66.2 & 61.6/68.7 \\
            $r=1/8$                   & | 55.6/65.9 & 58.0/66.3 & 61.6/68.7 \\
            $r=1/16$                  & | 53.3/65.7 & 56.9/66.2 & 61.6/68.7 \\
            \midrule
            Known $\mathbf{m}$        & | 57.3/66.0 & 59.0/66.9 & 61.4/68.7 \\
            \midrule
        \end{tabular} \\
    \textbf{(b) $\mathbf{m}$ correction ratio} (ACA/Acc).
    \end{tabular}
 
\end{tabular}
}
\end{table}

\section{Discussion}
\label{sec:conclusion}

The proposed SS-Text-U provides consistent gains across most datasets, as shown in Fig.~\ref{fig:per_dataset}. For specific tasks, e.g., NCT, the 1-shot performance is even comparable to that in large data regimes. However, we observe that \textit{all solvers} struggle in a few tasks, where they show moderate performance, e.g., mBRSET. To explore this issue, we use the Silhouette score~\cite{Silhouettes} ($S\in(-1, +1)$, where $1$ is perfect clustering and $0$ severe overlap) of the embeddings w.r.t. the \textit{true labels}. Interestingly, Fig~\ref{fig:ablation_studies}(d) shows a strong linear correlation ($\rho$) between the solver performance and $S$. This is not the case for zero-shot predictions, which also depend on non-accessible priors of the pre-training data~\cite{udandarao2024zeroshot}. In some of these difficult tasks, the zero-shot accuracy is exceptionally high. Also, their Silhouette scores closer to $S=0$ indicate that the embedding representations and/or label qualities do not suffice to enable well-performing few-shot classifiers. As future work, while we focused on delivering an efficient, well-principled solver using only feature embeddings (\textit{as the baseline linear probes}), other semi-supervised frameworks, e.g., multi-view augmentations and confidence filtering~\cite{fixmatch}, might be explored.

\bibliographystyle{splncs04}
\bibliography{refs}

\begin{thebibliography}{10}
\providecommand{\url}[1]{\texttt{#1}}
\providecommand{\urlprefix}{URL }
\providecommand{\doi}[1]{https://doi.org/#1}

\bibitem{boudiaf2020unifying}
Boudiaf, M., et~al.: A unifying mutual information view of metric learning: cross-entropy vs. pairwise losses. In: ECCV (2020)

\bibitem{CCRCC}
Brummer, O., et~al.: {Computational textural mapping harmonises sampling variation and reveals multidimensional histopathological fingerprints}. Br J Cancer  \textbf{129},  683–695 (2023)

\bibitem{Bustos2019PadChest:Reports}
Bustos, A., et~al.: {PadChest: A large chest x-ray image dataset with multi-label annotated reports}. Medical Image Analysis  \textbf{66} (2019)

\bibitem{chapelle2009semi}
Chapelle, O., Scholkopf, B., Zien, A.: Semi-supervised learning. IEEE Transactions on Neural Networks  \textbf{20}(3),  542--542 (2009)

\bibitem{covid1}
Chowdhury, M.E.H., et~al.: Can ai help in screening viral and covid-19 pneumonia? IEEE Access  \textbf{8},  132665--132676 (2020)

\bibitem{cuturi2013sinkhorn}
Cuturi, M.: Sinkhorn distances: Lightspeed computation of optimal transport. In: NeurIPS. vol.~26 (2013)

\bibitem{messidor}
Decencière, E., et~al.: Feedback on a publicly distributed image database: The messidor database. Image Analysis \& Stereology  \textbf{33},  231--234 (07 2014)

\bibitem{nih_lt}
Holste, G., et~al.: Long-tailed classification of thorax diseases on chest x-ray: A new benchmark study. In: DALI, MICCAI Workshops (2022)

\bibitem{lp24}
Huang, Y., et~al.: Lp++: A surprisingly strong linear probe for few-shot clip. In: CVPR (2024)

\bibitem{ikezogwo2023quiltm}
Ikezogwo, W.O., et~al.: Quilt-1m: One million image-text pairs for histopathology. In: NeurIPS Datasets and Benchmarks Track (2023)

\bibitem{irvin2019chexpert}
Irvin, J., et~al.: Chexpert: A large chest radiograph dataset with uncertainty labels and expert comparison. In: AAAI (2019)

\bibitem{fives}
Jin, K., et~al.: Fives: A fundus image dataset for artificial intelligence based vessel segmentation. Scientific Data  \textbf{9}, ~475 (08 2022)

\bibitem{kather2018100}
Kather, J.N., Halama, N., Marx, A.: 100,000 histological images of human colorectal cancer and healthy tissue. Zenodo10  \textbf{5281} (2018)

\bibitem{kriegsmann2022deep}
Kriegsmann, K., et~al.: Deep learning for the detection of anatomical tissue structures and neoplasms of the skin on scanned histopathological tissue sections. Frontiers in Oncology  \textbf{12},  1022967 (2022)

\bibitem{lee2013pseudo}
Lee, D.H., et~al.: Pseudo-label: The simple and efficient semi-supervised learning method for deep neural networks. In: Challenges in representation learning, ICML Workshops. vol.~3, p.~896 (2013)

\bibitem{lin2023crossmodal}
Lin, Z., et~al.: Multimodality helps unimodality: Cross-modal few-shot learning with multimodal models. In: CVPR (2023)

\bibitem{CONCH}
Lu, M.Y., et~al.: A visual-language foundation model for computational pathology. Nature Medicine  \textbf{30},  863–874 (2024)

\bibitem{metrics}
Maier-Hein, L., et~al.: Metrics reloaded: recommendations for image analysis validation. Nature Methods  \textbf{21} (2024)

\bibitem{shu2022tpt}
Manli, S., et~al.: Test-time prompt tuning for zero-shot generalization in vision-language models. In: NeurIPS (2022)

\bibitem{mBRSET}
Nakayama, L.F., et~al.: {mBRSET, a Mobile Brazilian Retinal Dataset}. PhysioNet  (2024)

\bibitem{mmac}
Qian, B., et~al.: A competition for the diagnosis of myopic maculopathy by artificial intelligence algorithms. JAMA ophthalmology  \textbf{142}(11),  1006--1015 (2024)

\bibitem{radford2021learning}
Radford, A., et~al.: Learning transferable visual models from natural language supervision. In: ICML. pp. 8748--8763 (2021)

\bibitem{covid2}
Rahman, T., et~al.: Exploring the effect of image enhancement techniques on covid-19 detection using chest x-ray images. Computers in Biology and Medicine  \textbf{132},  104319 (2021)

\bibitem{razaviyayn2013unified}
Razaviyayn, M., Hong, M., Luo, Z.Q.: A unified convergence analysis of block successive minimization methods for nonsmooth optimization. SIAM Journal on Optimization  \textbf{23}(2),  1126--1153 (2013)

\bibitem{Silhouettes}
Rousseeuw, P.: Silhouettes: a graphical aid to the interpretation and validation of cluster analysis. Journal of Computational and Applied Mathematics  \textbf{20}(1),  53–65 (1987)

\bibitem{shakeri2024few}
Shakeri, F., et~al.: Few-shot adaptation of medical vision-language models. In: MICCAI (2024)

\bibitem{tta_medvlm}
Shakeri, F., et~al.: Test-time adaptation of medical vision-language models. In: MedAGI, MICCAI Workshops (2025)

\bibitem{silva2020going}
Silva-Rodr{\'\i}guez, J., et~al.: Going deeper through the gleason scoring scale: An automatic end-to-end system for histology prostate grading and cribriform pattern detection. Computer methods and programs in biomedicine  \textbf{195},  105637 (2020)

\bibitem{clap24}
Silva-Rodr\'iguez, J., et~al.: A closer look at the few-shot adaptation of large vision-language models. In: CVPR (2024)

\bibitem{dlilp}
Silva-Rodríguez, J., Dolz, J., {Ben Ayed}, I.: A reality check of vision-language pre-training in radiology: Have we progressed using text? In: IPMI (2025)

\bibitem{sstext25}
Silva-Rodríguez, J., et~al.: Few-shot, now for real: Medical vlms adaptation without balanced sets or validation. In: MICCAI (2025)

\bibitem{FLAIR}
Silva-Rodríguez, J., et~al.: A foundation language-image model of the retina (flair): Encoding expert knowledge in text supervision. Medical Image Analysis  \textbf{99},  103357 (2025)

\bibitem{fca25}
Silva-Rodríguez, J., et~al.: Full conformal adaptation of medical vision-language models. In: IPMI (2025)

\bibitem{fixmatch}
Sohn, K., et~al.: Fixmatch: simplifying semi-supervised learning with consistency and confidence. In: NeurIPS (2020)

\bibitem{tseng2001convergence}
Tseng, P.: Convergence of a block coordinate descent method for nondifferentiable minimization. Journal of Optimization Theory and Applications  \textbf{109}(3),  475--494 (2001)

\bibitem{udandarao2024zeroshot}
Udandarao, V., et~al.: No "zero-shot" without exponential data: Pretraining concept frequency determines multimodal model performance. In: NeurIPS (2024)

\bibitem{ot_book}
Villani, C.: Optimal transport: old and new. Springer (2009)

\bibitem{nih}
Wang, X., et~al.: Chestx-ray8: Hospital-scale chest x-ray database and benchmarks on weakly-supervised classification and localization of common thorax diseases. In: CVPR (2017)

\bibitem{simpleshot}
Wang, Y., et~al.: Simpleshot: Revisiting nearest-neighbor classification for few-shot learning. In: arXiv preprint arXiv:1911.04623 (2019)

\bibitem{wright2015coordinate}
Wright, S.J.: Coordinate descent algorithms. Mathematical Programming  \textbf{151}(1),  3--34 (2015)

\bibitem{yu2023task}
Yu, T., Lu, Z., Jin, X., Chen, Z., Wang, X.: Task residual for tuning vision-language models. In: CVPR (2023)

\bibitem{zanella2024boosting}
Zanella, M., et~al.: Boosting vision-language models with transduction. In: NeurIPS (2024)

\bibitem{convirt}
Zhang, Y., et~al.: Contrastive learning of medical visual representations from paired images and text. In: MHLC (2022)

\bibitem{zhou2022coop}
Zhou, K., Yang, J., Loy, C.C., Liu, Z.: Learning to prompt for vision-language models. International Journal of Computer Vision  \textbf{130},  2337–2348 (2022)

\end{thebibliography}

\end{document}